\documentclass[runningheads]{llncs}

\usepackage[year=2026,ID=975]{accv}
\usepackage{accvabbrv}
\usepackage[pagebackref,breaklinks,colorlinks,citecolor=accvblue]{hyperref}
\usepackage{amsmath,amssymb}
\usepackage{booktabs}
\usepackage{multirow}
\usepackage{graphicx}
\usepackage{placeins}
\usepackage{orcidlink}

\begin{document}

\title{LAS-CLIP: A Lightweight Adapter Steering Approach for CLIP's Visual Encoder}
\titlerunning{LAS-CLIP}

\author{Anh-Khoa Dinh-Duc\inst{1, 2}\orcidlink{0009-0001-8569-1976} \and
Duc-Tai Dinh\inst{1, 2}\orcidlink{0009-0002-4958-8960} \and
Tam V. Nguyen\inst{3}\orcidlink{0000-0003-0236-7992} \and
Minh-Triet Tran\inst{1, 2}\orcidlink{0000-0003-3046-3041}}
\authorrunning{Dinh-Duc et al.}

\institute{
University of Science, Ho Chi Minh City, Vietnam \\
\and Viet Nam National University, Ho Chi Minh City, Vietnam\\
\and University of Dayton, USA \\
\email{ddakhoa@selab.hcmus.edu.vn, 23122013@student.hcmus.edu.vn, tamnguyen@udayton.edu, tmtriet@fit.hcmus.edu.vn} \\
\url{https://github.com/AnhKhoa585/lasclip}
}

\maketitle

\begin{abstract}
CLIP's visual encoder produces only global image representations, limiting its use in region-level tasks. Existing adaptations rely on visual prompting, input masking, or encoder fine-tuning, each compromising pre-trained representations. We propose LAS-CLIP, a Lightweight Adapter Steering approach that keeps every CLIP parameter frozen. A compact MaskAdapter generates per-head, per-layer attention biases from an input mask and injects them into the frozen self-attention layers, steering attention toward the target region. Crucially, because the backbone remains strictly untouched, LAS-CLIP seamlessly reverts to vanilla CLIP when no mask is provided, preserving its foundational zero-shot capabilities. With approximately 116K to 145K trainable parameters and 100K training samples on two T4 GPUs, LAS-CLIP achieves competitive or superior results compared to Alpha-CLIP on ImageNet-S zero-shot classification and RefCOCO referring expression comprehension, despite the latter fine-tuning its entire encoder on millions of samples. Qualitative analysis further confirms stronger representational fidelity under incorrect masks and in downstream generation.

\end{abstract}

\section{Introduction}
\label{sec:intro}
Contrastive Language-Image Pre-training (CLIP)~\cite{clip} has become a foundational building block for vision-language understanding. By training on hundreds of millions of image-text pairs, CLIP acquires a shared embedding space that enables strong zero-shot transfer across diverse visual recognition tasks. Subsequent work has scaled this paradigm further through larger datasets~\cite{align,openclip}, improved loss formulations~\cite{siglip}, and stronger visual backbones~\cite{evaclip}, solidifying contrastive vision-language models as a core component of modern visual pipelines.

Despite this progress, a fundamental limitation remains: CLIP's visual encoder produces a single global representation for the entire image. Many practical applications, including referring expression comprehension, open-vocabulary detection, and subject-driven image generation, require the model to focus on a specific region rather than the whole scene. Bridging this gap between image-level representations and region-level understanding is therefore an important research challenge.

Existing approaches to equip CLIP with region awareness can be broadly grouped into three strategies detailed in Section~\ref{subsec:rw_region}, each carrying notable drawbacks. For instance, visual prompting~\cite{redcircle,cpt} introduces a distribution gap by altering pixel content, input masking~\cite{maskadaptedclip,reclip} destroys contextual cues, and encoder modification~\cite{alphaclip} permanently alters the pre-trained weights while demanding large-scale data and substantial computational resources.

In this paper, we propose \textbf{LAS-CLIP}, a \textbf{L}ightweight \textbf{A}dapter \textbf{S}teering approach that adapts CLIP to region-level understanding while keeping every pre-trained parameter frozen. The core idea is a compact \textit{MaskAdapter} module that takes a binary mask as input and generates per-head, per-layer attention biases. These biases are injected directly into the multi-head self-attention layers of the frozen visual encoder, steering the model's focus toward the mask-specified region without modifying any original weight. Because the backbone remains untouched, LAS-CLIP seamlessly reverts to vanilla CLIP behavior when no mask is provided, a property that encoder modification methods cannot offer. An identity regularization term further controls the drift from CLIP's original representations.

Our contributions are as follows:
\begin{itemize}
\vspace{-2mm}
  \item We introduce LAS-CLIP, a parameter-efficient adapter that steers CLIP's visual attention toward user-specified regions through learned attention biases, requiring as few as {\raise.17ex\hbox{$\scriptstyle\sim$}}116K trainable parameters while keeping the entire backbone frozen.
  \item We design a gated token-mask interaction mechanism that produces attention biases dynamically conditioned on both the spatial mask and current visual tokens. Unlike static or uniform biases that force all attention heads to behave identically, our dynamic gating selectively modulates each head based on its specific semantic role, preserving CLIP's diverse feature representations while achieving precise spatial grounding.
  \item We demonstrate that LAS-CLIP achieves competitive or superior performance compared to Alpha-CLIP~\cite{alphaclip} on zero-shot classification and referring expression comprehension benchmarks, despite using only 100K training samples and two T4 GPUs. Qualitative analysis on attention visualization and subject-driven image generation further confirms that the frozen backbone retains stronger representational fidelity.
\end{itemize}

The remainder of this paper is organized as follows. Section~\ref{sec:related} reviews related work on CLIP adaptation and parameter-efficient fine-tuning. Section~\ref{sec:method} presents the proposed LAS-CLIP framework, including the MaskAdapter architecture, gated token-mask interaction, training objective, and data preparation pipeline. Section~\ref{sec:experiments} reports experimental results on zero-shot classification, referring expression comprehension, ablation studies, and qualitative analysis. Section~\ref{sec:conclusion} concludes the paper and discusses current limitations and future research directions.

\section{Related Work}
\label{sec:related}

\subsection{Vision-Language Contrastive Pre-training}
\label{subsec:rw_vlp}
CLIP~\cite{clip} and its successors~\cite{align,openclip,siglip,evaclip} have established contrastive vision-language pre-training as a foundational paradigm, yet all resulting models produce global image representations without the ability to attend to user-specified regions.

\subsection{Region-Aware CLIP Adaptation}
\label{subsec:rw_region}
Adapting CLIP to understand specific image regions has attracted considerable attention. Existing approaches can be broadly grouped into three strategies, namely, visual prompting, input masking, and encoder modification.

\paragraph{Visual Prompting.}
Shtedritski et al.~\cite{redcircle} showed that drawing a red circle around an object serves as an effective visual prompt for CLIP, and CPT~\cite{cpt} explored a similar strategy using colored fills. These approaches are training-free but modify pixel content, introducing a domain gap with CLIP's pre-training distribution. FALIP~\cite{falip} instead inserts foveal attention masks into CLIP's self-attention, though its bias is uniform across all heads, limiting fine-grained spatial steering.

\paragraph{Input Masking.}
MaskAdaptedCLIP~\cite{maskadaptedclip} fine-tunes CLIP on masked image regions with the background replaced by a solid color, but this destroys contextual information and degrades classification performance. ReCLIP~\cite{reclip} instead applies Gaussian blurring to the background, preserving more context while still operating at the input level. RegionCLIP~\cite{regionclip} pre-trains a region-text alignment objective on top of CLIP features for open-vocabulary detection.

\paragraph{Encoder Modification.}
Alpha-CLIP~\cite{alphaclip} augments CLIP's patch embedding layer with an auxiliary alpha channel and fine-tunes the visual encoder on millions of RGBA region-text pairs from GRIT~\cite{grit}. While effective, this approach permanently alters the pre-trained weights and demands substantial computational resources.

\subsection{Parameter-Efficient Fine-Tuning}
\label{subsec:rw_peft}
Adapting large pre-trained models without full fine-tuning has been widely studied through approaches such as bottleneck adapters~\cite{adapter}, bias-only tuning~\cite{bitfit}, and low-rank weight decomposition~\cite{lora}. In the vision-language domain, prompt-based methods~\cite{coop,cocoop,vpt} optimize learnable tokens on either the textual or visual branch of CLIP, while feature-level approaches~\cite{clipadapter,tipadapter} add lightweight residual layers on top of frozen features. These methods target primarily classification tasks and modify representations in a class-agnostic manner. Notably, none provides a native architectural mechanism to dynamically ingest spatial inputs, such as binary masks, and steer region-specific attention at the level of individual attention heads. This leaves a gap between parameter-efficient adaptation and fine-grained spatial grounding.

\section{Method}
\label{sec:method}

This section presents the proposed LAS-CLIP framework. We introduce the necessary preliminaries, describe the overall architecture, propose MaskAdapter for generating region-conditioned attention biases, and present the training objective and the data preparation pipeline used to train the lightweight adapter while keeping the CLIP backbone entirely frozen.

\subsection{Preliminary}
\label{subsec:preliminary}

\paragraph{CLIP.}
Contrastive Language-Image Pre-training (CLIP)~\cite{clip} learns aligned visual and textual representations through contrastive learning on large-scale image-text pairs. It consists of an image encoder $\mathbf{V}$ and a text encoder $\mathbf{T}$. Given an image $\mathbf{x}$ and a text description $c$, CLIP produces $\ell_2$-normalized embeddings $\mathbf{v} = \mathbf{V}(\mathbf{x})$ and $\mathbf{t} = \mathbf{T}(c)$, and measures their compatibility via cosine similarity $\mathbf{v}^\top \mathbf{t}$ since the embeddings are $\ell_2$-normalized. By training on millions of paired samples with a symmetric contrastive objective, CLIP acquires strong zero-shot transfer capability across diverse visual recognition tasks.

\paragraph{Vision Transformer.}
When CLIP adopts a Vision Transformer (ViT)~\cite{vit} as the image encoder $\mathbf{V}$, the input image $\mathbf{x} \in \mathbb{R}^{3 \times H \times W}$ is divided into a grid of non-overlapping patches, each of size $p \times p$. These patches are linearly projected into $D$-dimensional token embeddings, and a learnable \texttt{[CLS]} token is prepended, yielding a sequence of $N$ tokens, which includes the prepended \texttt{[CLS]} token. Within each transformer layer, multi-head self-attention (MSA) computes, for each attention head $i$, the query, key, and value projections $Q_i, K_i, V_i$ from the input tokens and produces the attention output as:
\begin{equation}
  \text{head}_i = \text{Softmax}\!\left(\frac{Q_i K_i^\top}{\sqrt{d}}\right) V_i,
  \label{eq:msa}
\end{equation}
where $d = D / H_a$ is the per-head dimension and $H_a$ is the number of attention heads. The outputs of all heads are concatenated and linearly projected. The final image representation is obtained from the \texttt{[CLS]} token at the last layer, followed by a projection head. This attention mechanism is the site where LAS-CLIP introduces learned biases to steer the model's focus toward mask-specified regions.

\paragraph{Alpha-CLIP.}
Alpha-CLIP~\cite{alphaclip} extends CLIP by augmenting its visual encoder with an auxiliary alpha channel. The input patch embedding convolution is expanded from 3 to 4 channels, accepting a binary mask as the fourth channel alongside the RGB image. The visual encoder is then fine-tuned with differential learning rates on millions of RGBA region-text pairs constructed from GRIT~\cite{grit}, applying a higher rate to the new alpha-channel parameters and a lower rate to the original weights, enabling the model to focus on mask-specified regions while preserving global recognition ability.

\subsection{Overview}
\label{subsec:overview}

\begin{figure}[t]
  \centering
  \includegraphics[width=\linewidth]{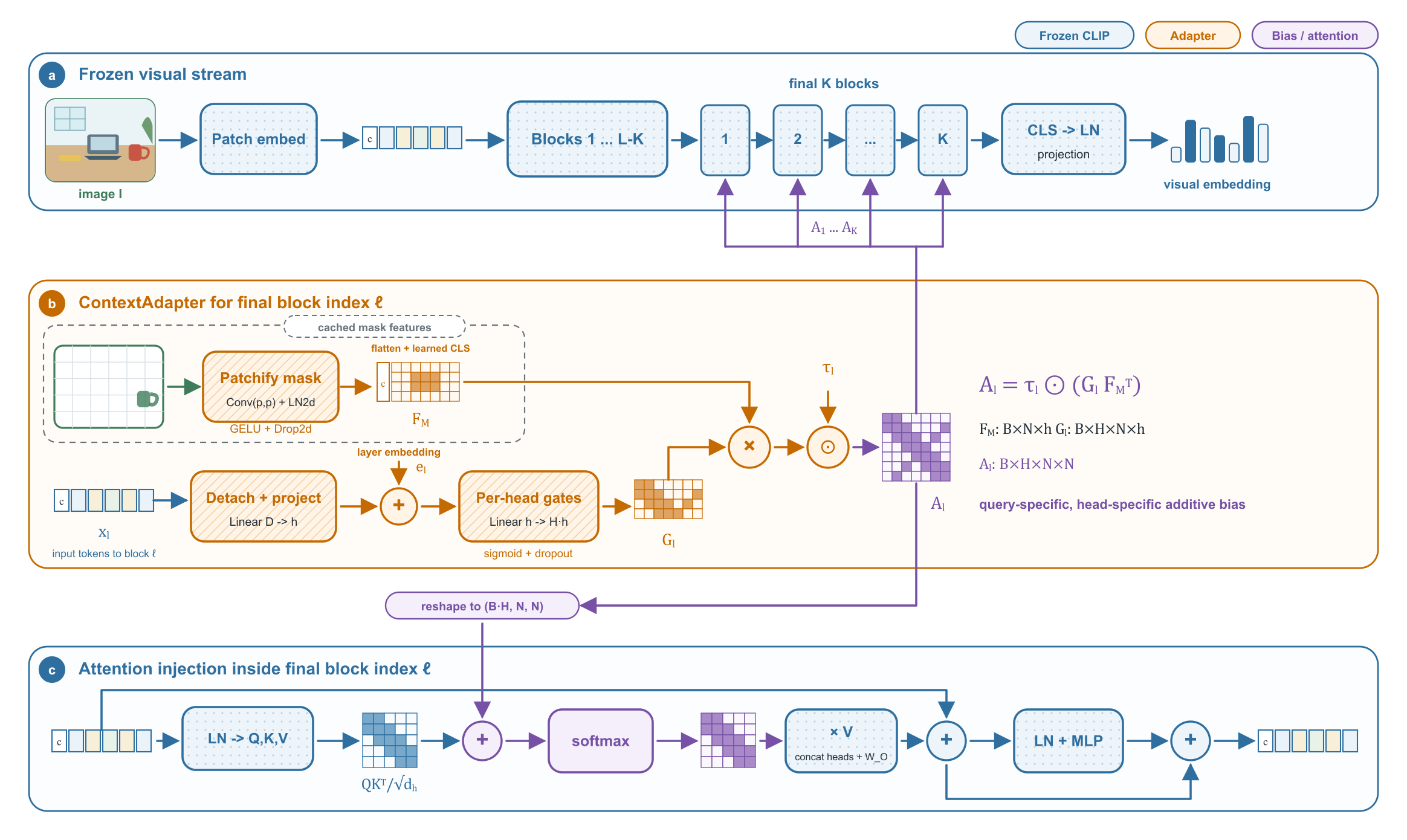}
  \caption{Overview of LAS-CLIP and the MaskAdapter architecture. The input image and a region mask are processed by the visual encoder and the MaskAdapter to produce region-aware features. The MaskAdapter consists of a mask encoder, a token encoder with layer conditioning, and a gated interaction module that computes attention biases to steer the frozen CLIP blocks.}
  \label{fig:overview}
\end{figure}

As illustrated in Figure~\ref{fig:overview}, given an input image $\mathbf{x} \in \mathbb{R}^{3 \times H \times W}$ and a binary mask $\mathbf{m} \in \mathbb{R}^{1 \times H \times W}$ indicating the region of interest, LAS-CLIP produces region-aware visual features while keeping the original CLIP weights entirely frozen. The key idea is to inject a lightweight \textit{MaskAdapter} module into the last $K$ transformer layers of CLIP's vision encoder $\mathbf{V}$. This module generates a per-head attention bias matrix $\mathbf{A}_{\text{bias}} \in \mathbb{R}^{B \times H_a \times N \times N}$, where $B$ is the batch size, $H_a$ is the number of attention heads, and $N$ is the number of visual tokens (including the \texttt{[CLS]} token). The bias is added directly to the attention logits $QK^\top$ within the frozen multi-head self-attention, steering the model to attend to mask-specified regions without modifying any pre-trained parameters. Identity regularization detailed in Section~\ref{subsec:training} further controls how far the adapted output may drift from the original representations.

In contrast to Alpha-CLIP~\cite{alphaclip}, which fine-tunes the visual encoder on millions of samples from GRIT~\cite{grit} using multiple A100-80G GPUs, LAS-CLIP is highly parameter-efficient and can be trained in a low-resource setting. Because all original weights remain strictly frozen, the model seamlessly reverts to vanilla CLIP when no mask is provided. This ensures zero degradation of the foundation model's original capabilities, a critical advantage over methods that permanently alter the backbone's architecture and weights.

Unlike prior approaches that apply a static, uniform bias shared across all attention layers and heads~\cite{falip}, LAS-CLIP introduces a dynamic, token-conditioned interaction. In standard Vision Transformers, different attention heads specialize in distinct visual patterns, ranging from local textures to global context. A uniform bias aggressively overrides this learned diversity. By generating per-head, per-layer biases through a gated mechanism, LAS-CLIP allows each head to independently decide the degree to which it should attend to the spatial mask, based on the current visual context. This formulation enables fine-grained and flexible steering of attention without collapsing the rich, multi-faceted representations pre-trained within CLIP.

\subsection{Gated Token-Mask Interaction}
\label{subsec:gated_interaction}

The MaskAdapter detailed in Figure~\ref{fig:overview} consists of three components. These include a mask encoder, a token encoder with layer conditioning, and a gated interaction module.

\paragraph{Mask Encoding.}
The binary mask $\mathbf{m}$ is processed by a convolutional branch that mirrors CLIP's patch embedding. A single 2D convolutional layer with kernel size and stride equal to the patch size $p$ converts $\mathbf{m}$ into patch-level mask features. These features are passed through layer normalization, GELU activation, and spatial dropout, yielding a feature map that is flattened into a sequence $\mathbf{F}_{\text{patch}} \in \mathbb{R}^{B \times P \times h}$, where $P$ is the number of patches and $h$ is the hidden dimension of the adapter. A learnable \texttt{[CLS]} mask token $\mathbf{f}_{\text{cls}} \in \mathbb{R}^{h}$ is prepended to this sequence, producing the final mask representation:
\begin{equation}
  \mathbf{F}_{\text{mask}} = [\mathbf{f}_{\text{cls}};\; \mathbf{F}_{\text{patch}}] \in \mathbb{R}^{B \times N \times h}.
  \label{eq:mask_feat}
\end{equation}
This encoding is computed once and shared across all $K$ adapter layers.

\paragraph{Token Encoding with Layer Conditioning.}
At each injected layer $k \in \{1, \dots, K\}$, the current visual tokens $\mathbf{X}^{(k)} \in \mathbb{R}^{B \times N \times D}$ from CLIP's transformer are detached from the backbone's computation graph and projected to the adapter's hidden dimension via a linear layer:
\begin{equation}
  \mathbf{f}_{\text{tok}}^{(k)} = \text{Linear}_D^h\bigl(\text{sg}(\mathbf{X}^{(k)})\bigr) + \mathbf{e}^{(k)},
  \label{eq:tok_feat}
\end{equation}
where $\text{sg}(\cdot)$ denotes the stop-gradient operator and $\mathbf{e}^{(k)} \in \mathbb{R}^{h}$ is a learnable layer embedding initialized to zero. The layer embedding conditions the adapter's behavior on the depth of the current layer, allowing different layers to produce distinct steering signals.

\paragraph{Gate Generation and Bias Computation.}
The token features $\mathbf{f}_{\text{tok}}^{(k)}$ are transformed into per-head gate vectors through a linear projection followed by sigmoid activation:
\begin{equation}
  \mathbf{G}^{(k)} = \sigma\bigl(\text{Linear}_h^{H_a \cdot h}(\mathbf{f}_{\text{tok}}^{(k)})\bigr) \in \mathbb{R}^{B \times H_a \times N \times h},
  \label{eq:gate}
\end{equation}
where $\sigma$ is the sigmoid function. Dropout is applied to the gate vectors during training. The attention bias is then computed by contracting the gate vectors with the mask features along the feature dimension $h$:
\begin{equation}
  A_{\text{bias},\,b\,a\,i\,j}^{(k)} = \tau_a^{(k)} \sum_{c=1}^{h} G_{b\,a\,i\,c}^{(k)}\; F_{\text{mask},\,b\,j\,c}, \quad \mathbf{A}_{\text{bias}}^{(k)} \in \mathbb{R}^{B \times H_a \times N \times N},
  \label{eq:bias}
\end{equation}
where $\tau_a^{(k)} \in \mathbb{R}$ is a learnable per-head, per-layer temperature parameter initialized to zero, with $a \in \{1, \dots, H_a\}$. This zero-initialization ensures that the adapter has no effect at the beginning of training.

\paragraph{Attention Steering.}
The computed bias is reshaped to $(B \cdot H_a, N, N)$ and added to the standard attention logits inside CLIP's multi-head self-attention:
\begin{equation}
  \text{Attn}_i(Q_i, K_i, V_i) = \text{Softmax}\!\left(\frac{Q_i K_i^\top}{\sqrt{d}} + \mathbf{A}_{\text{bias},i}^{(k)}\right) V_i,
  \label{eq:attn}
\end{equation}
where $d$ is the per-head dimension. This formulation allows each attention head to independently modulate its focus based on both the spatial mask and the current token semantics. Specifically, the sigmoid-activated gate (Eq.~\ref{eq:gate}) acts as an adaptive filter that prevents the mask signal from overwhelming the heads responsible for capturing global context, thereby preserving the representational diversity of the frozen backbone.

\subsection{Training Objective}
\label{subsec:training}
As outlined in Section~\ref{subsec:overview}, only the MaskAdapter parameters are optimized, enabling efficient training on commodity hardware with a batch size of 128.

\paragraph{Contrastive Loss.}
The primary training signal is a symmetric contrastive loss following the standard CLIP formulation. Given a batch of $B$ image-mask-text triplets $\{(\mathbf{x}_i, \mathbf{m}_i, t_i)\}_{i=1}^{B}$, we compute region-aware image features $\mathbf{v}_i = \mathbf{V}(\mathbf{x}_i, \mathbf{m}_i)$ and text features $\mathbf{t}_i = \mathbf{T}(t_i)$, both $\ell_2$-normalized. The contrastive loss is:
\begin{equation}
  \mathcal{L}_{\text{con}} = \frac{1}{2}\bigl(\text{CE}(\mathbf{S}, \mathbf{y}) + \text{CE}(\mathbf{S}^\top, \mathbf{y})\bigr),
  \label{eq:contrastive}
\end{equation}
where $\mathbf{S}_{ij} = \lambda \cdot \mathbf{v}_i^\top \mathbf{t}_j$ is the scaled similarity matrix, $\lambda$ is the learnable logit scale, CE is the cross-entropy loss, and $\mathbf{y}$ is the identity label vector.

\paragraph{Identity Regularization.}
To control the degree to which the adapter reshapes CLIP's pre-trained representations, we introduce a configurable identity regularization term. Let $\hat{\mathbf{v}}_i$ denote the feature produced by the vanilla (unmodified) CLIP encoder for the same image $\mathbf{x}_i$. The regularization loss encourages the adapter output to remain close to the original representation:
\begin{equation}
  \mathcal{L}_{\text{id}} = 1 - \frac{1}{B} \sum_{i=1}^{B} \cos(\mathbf{v}_i, \hat{\mathbf{v}}_i),
  \label{eq:identity}
\end{equation}
where $\cos(\cdot, \cdot)$ denotes cosine similarity. The total training objective is:
\begin{equation}
  \mathcal{L} = \mathcal{L}_{\text{con}} + \alpha \cdot \mathcal{L}_{\text{id}},
  \label{eq:total_loss}
\end{equation}
where $\alpha$ is a balancing weight.

\paragraph{Full-Image Augmentation.}
During training, with a small probability $p_{\text{aug}}$, the mask is replaced with an all-ones mask covering the entire image, and the text is replaced with the original image caption. This augmentation strategy prevents the model from forgetting its holistic image understanding capability. Additional optimization details and hyperparameter configurations are provided in the supplementary material.

\subsection{Data Preparation}
\label{subsec:data}
Training LAS-CLIP requires paired image-mask-text data. We adopt the grounding data pipeline introduced by Alpha-CLIP~\cite{alphaclip}, which constructs region-text pairs from the GRIT dataset~\cite{grit}. GRIT provides images annotated with bounding boxes and referring expressions. Alpha-CLIP uses the Segment Anything Model, or SAM~\cite{sam}, to convert these bounding boxes into segmentation masks. In our work, we replace SAM with YOLOE~\cite{yoloe}. While SAM provides highly precise boundaries, its inference is computationally expensive and excessively slow on standard resource-constrained hardware such as T4 GPUs. By utilizing YOLOE, we establish a highly efficient data pipeline that is designed to be easily scaled up to millions of images in the future with minimal computational cost. More importantly, utilizing YOLOE demonstrates the robustness of LAS-CLIP: we show that our model can effectively learn accurate attention biases even when trained on noisy or imperfect mask annotations, eliminating the strict dependency on expensive, pixel-perfect masks.

For each annotated bounding box, YOLOE predicts a corresponding segmentation mask. A common challenge for any box-prompted segmentation model, including SAM, is that when a bounding box tightly encloses multiple objects, the model may occasionally segment an unintended instance, resulting in a mask that poorly aligns with the target region. To address this general ambiguity, as well as the natural noise in high-speed segmentation pipelines, we apply an IoU-based fallback. Specifically, if the predicted mask has an Intersection-over-Union, or IoU, with the bounding box below a threshold of 0.30, which was empirically chosen based on visual inspection for optimal training stability, we fall back to using the full bounding box region as the mask.

\paragraph{Quality Filtering.}
We apply two filtering criteria to ensure data quality. The mask area ratio relative to the image must lie within $[0.01, 0.95]$, discarding regions that are too small or cover nearly the entire image. In addition, referring expressions shorter than 3 characters are removed. After filtering, we retain 100K high-quality training samples. Further details on the mask generation process and fallback statistics can be found in the supplementary material.

\section{Experiments}
\label{sec:experiments}
To comprehensively validate the effectiveness of our lightweight steering mechanism, we evaluate LAS-CLIP across three distinct paradigms of region-aware understanding. We assess zero-shot image classification on ImageNet-S~\cite{imagenets} to verify that the model retains robust global semantic recognition. We then evaluate zero-shot referring expression comprehension across three demanding datasets, RefCOCO~\cite{refcoco}, RefCOCO+~\cite{refcoco}, and RefCOCOg~\cite{refcocog}, to strictly test fine-grained spatial and relational reasoning. We also demonstrate the plug-and-play generative utility of our frozen encoder in a downstream subject-driven image generation task using BLIP-Diffusion~\cite{blipdiffusion}. All models are trained under the low-resource protocol described in Section~\ref{subsec:data}. The MaskAdapter is injected into the last $K\!=\!3$ transformer layers of CLIP's vision encoder unless stated otherwise. Full training hyperparameters are provided in the supplementary material.

\subsection{Experimental Setup}
\label{subsec:setup}

\paragraph{Datasets.}
For zero-shot classification, we adopt the ImageNet-S~\cite{imagenets} dataset, which contains 919 classes selected from ImageNet-1k~\cite{imagenet} with semantic segmentation annotations. The image-level segmentation masks serve as the region input, and the mean of per-class accuracy is used as the evaluation metric. For zero-shot REC, we evaluate on RefCOCO~\cite{refcoco}, RefCOCO+~\cite{refcoco}, and RefCOCOg~\cite{refcocog}, which are built on the MS COCO~\cite{mscoco} dataset. Following Alpha-CLIP~\cite{alphaclip}, we use object proposals from a pretrained MAttNet~\cite{mattnet} detector and obtain masks for each proposal. The original image together with the corresponding mask is fed to LAS-CLIP, and the proposal with the highest similarity to the referring expression is selected. We report accuracy using an ensemble of ViT-B/16 and ViT-L/14 backbones on the standard val and test splits.

\paragraph{Baselines.}
We compare LAS-CLIP against the original CLIP~\cite{clip}, MaskAdaptedCLIP~\cite{maskadaptedclip}, Red Circle~\cite{redcircle}, and Alpha-CLIP~\cite{alphaclip}. For the REC task, we additionally include CPT~\cite{cpt} and ReCLIP~\cite{reclip}. Both ViT-B/16 and ViT-L/14 backbones are evaluated.

\subsection{Zero-shot Image Classification}
\label{subsec:result_cls}

Table~\ref{tab:imagenets} presents the zero-shot classification results on ImageNet-S. All attention-bias methods improve over vanilla CLIP~\cite{clip}, while MaskAdaptedCLIP~\cite{maskadaptedclip} and Red Circle~\cite{redcircle} underperform it, confirming that learned biases outperform pixel-level modifications. FALIP~\cite{falip} achieves moderate gains through uniform biases, but LAS-CLIP surpasses both FALIP and Alpha-CLIP~\cite{alphaclip} on ViT-B/16 and on ViT-L/14 Top-1, demonstrating the benefit of token-conditioned, per-head steering. We adopt $K\!=\!3$ as the default configuration as analyzed in Section~\ref{subsec:ablation}.

\begin{table}[h]
\centering
\caption{Zero-shot classification on ImageNet-S validation set. Mean per-class accuracy in percentage is reported. The best results are in \textbf{bold} and the second best results are \underline{underlined}.}
\label{tab:imagenets}
\setlength{\tabcolsep}{7pt}
\begin{tabular}{l cc @{\hspace{1.5em}} cc}
\toprule
\multirow{2}{*}{Method} & \multicolumn{2}{c}{B/16} & \multicolumn{2}{c}{L/14} \\
\cmidrule(lr){2-3} \cmidrule(lr){4-5}
 & Top-1 & Top-5 & Top-1 & Top-5 \\
\midrule
Original CLIP~\cite{clip}       & 66.5 & 88.9 & 73.5 & 91.6 \\
MaskAdaptedCLIP~\cite{maskadaptedclip} & 57.9 & 79.1 & 63.5 & 86.3 \\
Red Circle~\cite{redcircle}     & 65.4 & 88.7 & 73.4 & 92.1 \\
FALIP~\cite{falip}              & 68.2 & 89.5 & 74.9 & 91.9 \\
Alpha-CLIP~\cite{alphaclip}     & \underline{68.9} & \underline{90.5} & \underline{77.4} & \textbf{94.5} \\
\midrule
LAS-CLIP (3 layers)             & \textbf{69.9} & \textbf{90.9} & \textbf{77.6} & \underline{94.0} \\
\bottomrule
\end{tabular}
\end{table}

\subsection{Zero-shot Referring Expression Comprehension}
\label{subsec:result_rec}

Table~\ref{tab:refcoco} presents the zero-shot REC results. LAS-CLIP surpasses all baselines on RefCOCO and RefCOCO+~\cite{refcoco}, with notable gains on the testA splits. On RefCOCOg~\cite{refcocog}, LAS-CLIP leads on val while Alpha-CLIP~\cite{alphaclip} retains a slight edge on test. FALIP~\cite{falip} trails both methods by a large margin, indicating that uniform attention biases without token conditioning are insufficient for fine-grained spatial reasoning.

\begin{table}[!h]
\centering
\caption{Zero-shot referring expression comprehension accuracy in percentage on RefCOCO, RefCOCO+, and RefCOCOg. Following our primary baseline Alpha-CLIP, LAS-CLIP utilizes an ensemble of ViT-B/16 and ViT-L/14 backbones to ensure a strictly fair comparison, FALIP also uses the same setting. Note that prior baselines also report ensemble results using their strongest respective configurations (e.g., ReCLIP ensembles RN50x16 and ViT-B/32, Red Circle ensembles RN50x16 and ViT-L/14@336px). The best results are in \textbf{bold} and the second best results are \underline{underlined}.}
\label{tab:refcoco}
\setlength{\tabcolsep}{4pt}
\begin{tabular}{l ccc ccc cc}
\toprule
\multirow{2}{*}{Method} & \multicolumn{3}{c}{RefCOCO} & \multicolumn{3}{c}{RefCOCO+} & \multicolumn{2}{c}{RefCOCOg} \\
\cmidrule(lr){2-4} \cmidrule(lr){5-7} \cmidrule(lr){8-9}
 & val & testA & testB & val & testA & testB & val & test \\
\midrule
CPT~\cite{cpt}               & 32.2 & 36.1 & 30.3 & 31.9 & 35.2 & 28.8 & 36.7 & 36.5 \\
ReCLIP~\cite{reclip}          & 45.8 & 46.1 & 47.1 & 47.9 & 50.1 & 45.1 & 59.3 & 59.0 \\
Red Circle~\cite{redcircle}   & 49.8 & 58.6 & 39.9 & 55.3 & \underline{63.9} & 45.4 & 59.4 & 58.9 \\
FALIP~\cite{falip}            & 43.8 & 45.6 & 42.4 & 44.1 & 46.5 & 40.0 & 48.8 & 48.7 \\
Alpha-CLIP~\cite{alphaclip}   & \underline{55.7} & \underline{61.1} & \textbf{50.3} & \underline{55.6} & 62.7 & \underline{46.4} & \underline{61.2} & \textbf{62.0} \\
\midrule
LAS-CLIP           & \textbf{57.5} & \textbf{64.1} & \underline{50.3} & \textbf{57.4} & \textbf{66.0} & \textbf{46.6} & \textbf{61.7} & \underline{61.7} \\
\bottomrule
\end{tabular}
\end{table}

\subsection{Ablation Study}
\label{subsec:ablation}

We ablate the number of adapter layers, identity regularization, and adapter architecture. For ImageNet-S, we report per-backbone results. For RefCOCO ablations on $K$ and $\mathcal{L}_{\text{id}}$, we follow the ensemble protocol from Section~\ref{subsec:result_rec}.

\paragraph{Number of Adapter Layers.}
Table~\ref{tab:ablation_layers} shows the effect of varying $K$. Fewer layers preserve classification accuracy, while more layers improve grounding up to $K\!=\!3$, which we adopt as the default.

\begin{table}[h]
\centering
\caption{Effect of the number of adapter layers $K$ on ImageNet-S top-1 accuracy (ViT-B/16 and ViT-L/14) and RefCOCO accuracy (ViT-B/16 + ViT-L/14 ensemble) in percentage. The best results are in \textbf{bold} and the second best results are \underline{underlined}.}
\label{tab:ablation_layers}
\setlength{\tabcolsep}{4pt}
\begin{tabular}{l c c @{\hspace{1.5em}} c c c}
\toprule
\multirow{2}{*}{Layers} & \multicolumn{2}{c}{ImageNet-S Top-1} & \multicolumn{3}{c}{RefCOCO} \\
\cmidrule(lr){2-3} \cmidrule(lr){4-6}
 & \makebox[3.5em][c]{B/16} & \makebox[3.5em][c]{L/14} & \makebox[3.5em][c]{val} & \makebox[3.5em][c]{testA} & \makebox[3.5em][c]{testB} \\
\midrule
$K\!=\!1$ & \textbf{70.7} & \textbf{78.1} & 54.9 & 60.9 & 49.6 \\
$K\!=\!2$ & \underline{70.0} & \underline{77.7} & 55.8 & 62.2 & \underline{50.3} \\
$K\!=\!3$ & 69.9 & 77.6 & \textbf{57.5} & \underline{64.1} & \underline{50.3} \\
$K\!=\!4$ & 69.6 & 76.6 & \underline{57.3} & \textbf{64.3} & \textbf{50.3} \\
\bottomrule
\end{tabular}

\end{table}

\paragraph{Effect of Identity Regularization.}
Table~\ref{tab:ablation_idreg} compares training with $K\!=\!4$ layers with and without $\mathcal{L}_{\text{id}}$. Removing it drops classification accuracy but improves grounding, as the adapter can reshape attention more aggressively. The weight $\alpha$ thus serves as a tunable knob to balance global semantics and spatial grounding without permanently altering the frozen backbone.

\begin{table}[h]
\centering
\caption{Effect of identity regularization on ImageNet-S top-1 accuracy (per-backbone) and RefCOCO accuracy (ViT-B/16 + ViT-L/14 ensemble) in percentage, using $K\!=\!4$ layers. Best results are in \textbf{bold}.}
\label{tab:ablation_idreg}
\setlength{\tabcolsep}{4pt}
\begin{tabular}{l c c @{\hspace{1.5em}} c c c}
\toprule
\multirow{2}{*}{$\mathcal{L}_{\text{id}}$} & \multicolumn{2}{c}{ImageNet-S Top-1} & \multicolumn{3}{c}{RefCOCO} \\
\cmidrule(lr){2-3} \cmidrule(lr){4-6}
 & \makebox[3.5em][c]{B/16} & \makebox[3.5em][c]{L/14} & \makebox[3.5em][c]{val} & \makebox[3.5em][c]{testA} & \makebox[3.5em][c]{testB} \\
\midrule
w/ & \textbf{69.6} & \textbf{76.6} & 57.3 & 64.3 & 50.3 \\
w/o & 66.4 & 76.1 & \textbf{58.2} & \textbf{65.2} & \textbf{50.6} \\
\bottomrule
\end{tabular}
\end{table}

\paragraph{Adapter Architecture.}
Table~\ref{tab:ablation_arch} isolates each architectural component using ViT-B/16. ``CLS-only'' restricts token conditioning to the \texttt{[CLS]} token, ``Head-agnostic'' shares a single bias across all heads, and ``No token'' removes token conditioning entirely. The full design achieves the best grounding accuracy across all splits.

\begin{table}[h]
\centering
\caption{Adapter architecture ablation on ViT-B/16. Best in \textbf{bold}.}
\label{tab:ablation_arch}
\setlength{\tabcolsep}{4pt}
\begin{tabular}{l cc ccc}
\toprule
\multirow{2}{*}{Variant} & \multicolumn{2}{c}{ImageNet-S} & \multicolumn{3}{c}{RefCOCO} \\
\cmidrule(lr){2-3} \cmidrule(lr){4-6}
 & Top-1 & Top-5 & val & testA & testB \\
\midrule
LAS-CLIP (default)  & 69.9 & 90.9 & \textbf{54.7} & \textbf{60.7} & \textbf{48.7} \\
CLS-only            & 69.9 & \textbf{91.0} & 53.5 & 60.2 & 48.1 \\
Head-agnostic       & \textbf{70.4} & \textbf{91.0} & 52.2 & 58.3 & 47.5 \\
No token cond.      & 69.9 & 90.8 & 52.4 & 58.7 & 48.1 \\
\bottomrule
\end{tabular}
\end{table}

Additional experiments in the supplementary material examine mask robustness under controlled perturbations, the effect of replacing YOLOE with SAM for training mask generation, and inference efficiency across batch sizes. These results confirm that LAS-CLIP degrades gracefully under imperfect masks, that the observed gains stem from the adapter design rather than the data pipeline, and that the adapter introduces comparable latency to Alpha-CLIP while consuming higher GPU memory.

\subsection{Qualitative Analysis}
\label{subsec:qualitative}

To better understand how each model attends to the specified region, we visualize the attention heatmaps using the text-based decomposition method proposed by Gandelsman et al.~\cite{clipdecomp}. This approach decomposes CLIP's image representation across individual patches, revealing where the model focuses given a text query. For these visualizations, we use the ViT-L/14 backbone, with LAS-CLIP operating under the 3-layer adapter configuration. We compare vanilla CLIP, Alpha-CLIP~\cite{alphaclip}, and LAS-CLIP. Vanilla CLIP takes only the original image as input. For Alpha-CLIP and LAS-CLIP, which receive both the original image and a mask, we evaluate under two conditions: with the correct mask and with an incorrect mask.

\begin{figure}[!t]
  \centering
  \includegraphics[width=\linewidth]{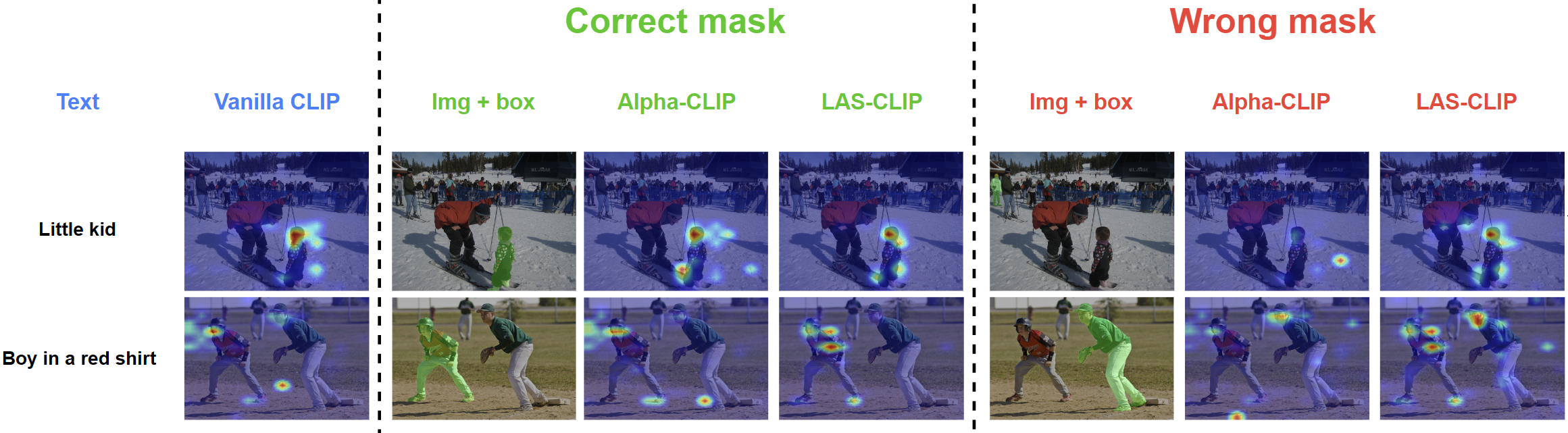}
  \caption{Attention heatmaps visualized via text-based decomposition~\cite{clipdecomp}. Each row shows a different text query. The left group shows results using correct masks, and the right group shows results using incorrect masks. LAS-CLIP maintains better text-aligned attention under incorrect mask inputs compared to Alpha-CLIP.}
  \label{fig:heatmap}
\end{figure}

Figure~\ref{fig:heatmap} visualizes representative examples. Under the correct mask condition, both Alpha-CLIP and LAS-CLIP successfully shift their attention toward the target region. However, the critical difference emerges under the wrong mask condition. When a mismatched mask is provided, Alpha-CLIP continues to attend strongly to the mask-indicated region even though it does not correspond to the text query, demonstrating that its fine-tuned encoder has become overly dependent on the mask input. In contrast, LAS-CLIP maintains attention patterns that remain closer to the text-relevant region despite the incorrect mask, suggesting that the frozen backbone preserves more of CLIP's original text-image alignment. Additional qualitative examples are provided in the supplementary material.

Beyond attention analysis, we evaluate region-aware CLIP variants as visual backbones for a downstream generative task. Following Alpha-CLIP~\cite{alphaclip}, we adapt BLIP-Diffusion~\cite{blipdiffusion} by replacing its default CLIP encoder with the evaluated variants. For baseline CLIP, subject features are extracted from the original reference image alone. For Alpha-CLIP and LAS-CLIP, we feed both the reference image and its segmentation mask directly into the visual backbone. To isolate the impact of visual representations, the text prompt is left blank, forcing the models to generate the subject solely from their visual inputs. This directly tests how faithfully each method captures the target subject. All methods use the ViT-L/14 backbone.

\begin{figure}[h!]
  \centering
  \includegraphics[width=0.6\linewidth]{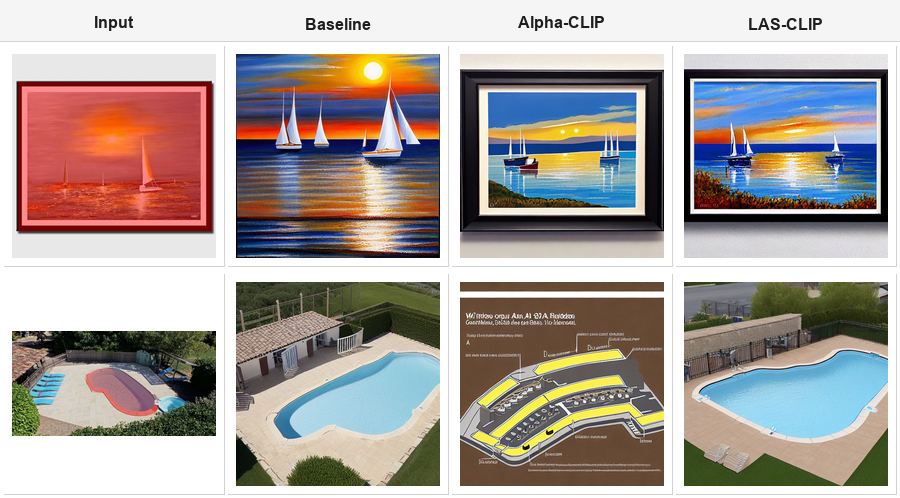}
  \caption{Subject-driven image generation using BLIP-Diffusion~\cite{blipdiffusion} with different CLIP backbones (ViT-L/14). Each row shows an input subject with its mask and the outputs generated by the baseline CLIP, Alpha-CLIP~\cite{alphaclip}, and LAS-CLIP. LAS-CLIP better preserves the visual identity of the input subject.}
  \label{fig:blipdiff}
\end{figure}

Figure~\ref{fig:blipdiff} presents representative results. The baseline CLIP, lacking any region-awareness mechanism, often generates images that deviate from the input subject in shape, color, or structural detail. Alpha-CLIP~\cite{alphaclip} improves subject fidelity in some cases, yet its fine-tuned encoder occasionally distorts subtle visual attributes or introduces artifacts that do not match the original subject. LAS-CLIP produces outputs that more closely preserve the visual identity of the input subject, maintaining consistent color, texture, and structural characteristics. This behavior aligns with the frozen backbone design of LAS-CLIP: because the pre-trained CLIP weights remain unmodified, the encoder retains its original capacity for fine-grained visual representation, and the MaskAdapter steers attention toward the masked region without degrading the underlying feature quality. Additional subject-driven generation examples are provided in the supplementary material.

\section{Conclusion}
\label{sec:conclusion}
We have presented LAS-CLIP, a parameter-efficient approach that injects a lightweight MaskAdapter into CLIP's frozen visual encoder, generating per-head, per-layer attention biases to steer attention toward user-specified regions without modifying any pre-trained weight. Because the backbone remains untouched, LAS-CLIP seamlessly reverts to vanilla CLIP when no mask is provided. Experiments on ImageNet-S, RefCOCO, RefCOCO+, and RefCOCOg demonstrate competitive or superior performance compared to full encoder fine-tuning methods, with a fraction of the trainable parameters and training data. Qualitative analysis further confirms that the frozen backbone retains stronger representational fidelity for downstream tasks.

Current limitations include the reliance on pre-computed segmentation masks and a higher GPU memory footprint due to dense per-head attention bias matrices. Future work will explore task-conditioned adaptive balancing, more flexible region specifications, and integration into dense prediction frameworks for open-vocabulary detection and semantic segmentation.

\subsubsection*{Acknowledgements}
This research is partially supported by the research funding from the Faculty of Information Technology, University of Science, VNU-HCM, Vietnam.

\clearpage

\begin{center}
\Large \textbf{LAS-CLIP: Supplementary Material}
\end{center}
\vspace{1cm}

\setcounter{section}{0}
\renewcommand{\thesection}{\Alph{section}}
\setcounter{figure}{0}
\renewcommand{\thefigure}{S\arabic{figure}}
\setcounter{table}{0}
\renewcommand{\thetable}{S\arabic{table}}
\setcounter{equation}{0}
\renewcommand{\theequation}{S\arabic{equation}}

\section{Implementation Details}
\label{sec:appendix_impl}
We provide additional implementation details to facilitate reproducibility.

\paragraph{Architecture.}
We initialize from the publicly available OpenAI CLIP pretrained weights and evaluate LAS-CLIP on two backbone variants. These variants are ViT-B/16 with $D\!=\!768$, $H_a\!=\!12$, $L\!=\!12$ and ViT-L/14 with $D\!=\!1024$, $H_a\!=\!16$, $L\!=\!24$. The MaskAdapter uses a hidden dimension of $h = 64$ for all configurations, resulting in only 116,004 trainable parameters for ViT-B/16 and 145,200 for ViT-L/14. Dropout with rate 0.2 is applied in both the mask encoder branch and the gate generator.

\paragraph{Optimization.}
We use AdamW~\cite{adamw} with a learning rate of $2 \times 10^{-3}$ for the adapter parameters. The logit scale $\lambda$ is trained with a separate learning rate of $1 \times 10^{-5}$. Weight decay is set to $2 \times 10^{-4}$. Training runs for 5 epochs with a OneCycleLR~\cite{onecyclelr} schedule and a 10\% linear warmup phase. To ensure stable contrastive learning, we apply 16 gradient accumulation steps to a base batch size of 128, yielding an effective batch size of 2048. Gradient clipping is applied at $\|\mathbf{g}\|_{\max} = 1.0$.

\paragraph{Training Cost.}
Thanks to the compact adapter design and the small number of trainable parameters, the entire training procedure is highly efficient. On two NVIDIA T4 GPUs, training the ViT-B/16 variant completes in approximately 2 hours, while the larger ViT-L/14 variant requires roughly 6 hours. This low computational overhead makes LAS-CLIP straightforward to reproduce on widely accessible, consumer-grade hardware without requiring high-end GPU clusters.

\paragraph{Loss Configuration.}
The identity regularization weight is $\alpha = 0.5$. The full-image augmentation probability is $p_{\text{aug}} = 0.1$.

\paragraph{Mask Generation Details.}
We specifically use the YOLOE-26x-seg model for mask prediction. During this process, 12,995 out of the 100,000 training samples fell back to the bounding box mask due to the 0.30 IoU threshold, successfully filtering out ambiguous box-prompt cases to ensure optimal training stability as discussed in Section 3.5 of the main paper.

\paragraph{Zero-shot Classification Protocol.}
We follow the standard CLIP zero-shot evaluation protocol using the 80 prompt templates provided by Radford et al.~\cite{clip}. For each class, text features are computed for all 80 templates and averaged to produce a single class embedding. The cosine similarity between the image feature and each averaged class embedding determines the prediction.

\paragraph{Zero-shot REC Evaluation Protocol.}
Following Alpha-CLIP~\cite{alphaclip}, we use object proposals predicted by a pretrained MAttNet~\cite{mattnet} detector. For each proposal, a segmentation mask is obtained using SAM~\cite{sam} with the proposal bounding box as the box prompt. The original image and the corresponding mask are fed to LAS-CLIP to produce a region-aware image feature. We also apply Gaussian blurring with standard deviation $\sigma = 100$ to the original image and compute a second feature from the blurred input paired with the same mask, following the preprocessing protocol of ReCLIP~\cite{reclip} and Alpha-CLIP~\cite{alphaclip}. The similarity scores from the original and blurred inputs are summed for each proposal, and the proposal with the highest total cosine similarity to the referring expression text feature is selected as the prediction.

For the backbone ensemble reported in the main paper, we compute similarity scores independently from both ViT-B/16 and ViT-L/14 backbones. The final score for each proposal is the sum of scores across both backbones, and the proposal with the highest combined score is selected.

\section{Additional Qualitative Results}
\label{sec:appendix_qual}

\begin{figure}[!t]
  \centering
  \includegraphics[width=\linewidth]{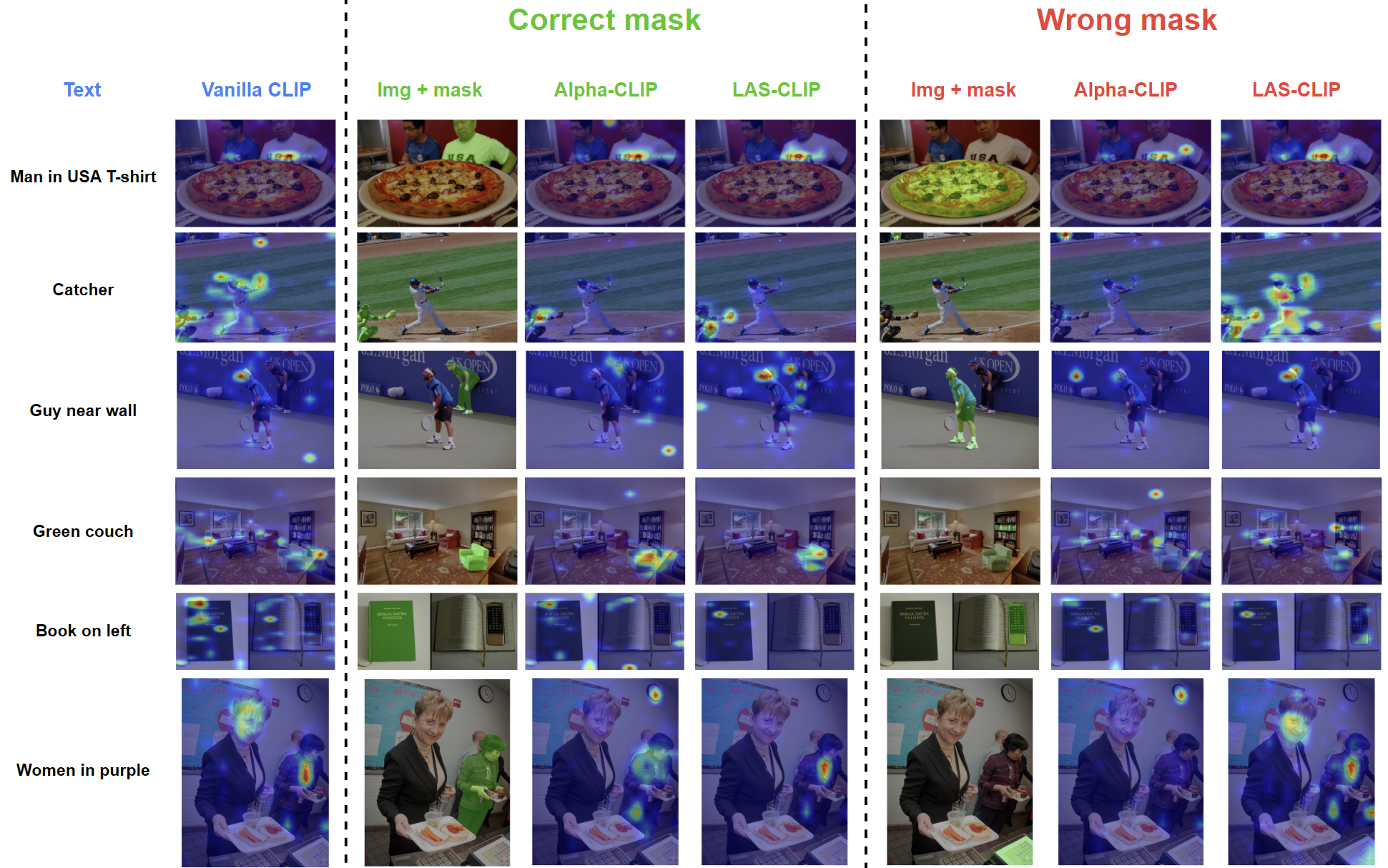}
  \caption{Additional attention heatmap visualizations. Each row shows a different text query. The left group shows results using correct masks, and the right group shows results using incorrect masks.}
  \label{fig:heatmap_more}
\end{figure}

Figure~\ref{fig:heatmap_more} provides additional attention heatmap visualizations beyond those shown in the main paper. Under the correct mask condition, all region-aware methods successfully concentrate attention on the target object. The key distinctions emerge under the wrong mask condition. For example, in the ``Man in USA T-shirt'' row, Alpha-CLIP shifts its attention to the pizza region indicated by the incorrect mask, whereas LAS-CLIP tends to maintain focused activation on the person. A similar trend is visible in the ``Book on left'' row, where Alpha-CLIP attends heavily to the wrong mask region while LAS-CLIP maintains query-relevant attention. Across these examples, LAS-CLIP appears to degrade more gracefully under incorrect spatial guidance. This supports the observation in the main paper that the frozen backbone helps preserve the original text-image alignment compared to Alpha-CLIP's fine-tuned encoder. We selected these cases to represent diverse challenges in object relationship and scale, though the degree of improvement varies across individual samples.

\section{Additional Subject-Driven Generation Results}
\label{sec:appendix_blipdiff}

\begin{figure}[htbp]
  \centering
  \includegraphics[width=\linewidth]{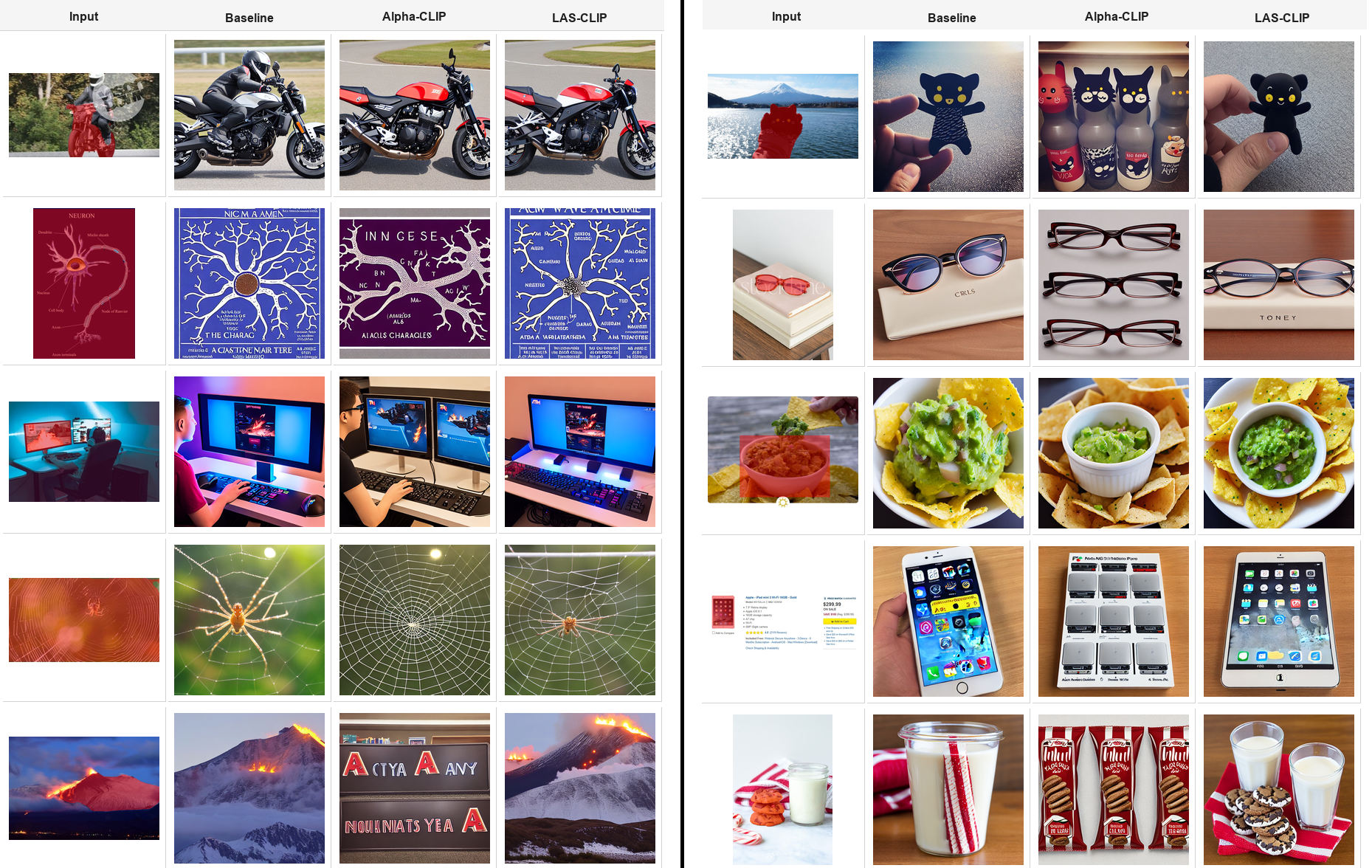}
  \caption{Additional subject-driven generation results using BLIP-Diffusion~\cite{blipdiffusion}. Each row shows an input subject, its mask, and the generated outputs.}
  \label{fig:blipdiff_more}
\end{figure}

Figure~\ref{fig:blipdiff_more} provides extended generation results using BLIP-Diffusion~\cite{blipdiffusion} beyond those shown in the main paper. Across the evaluated subjects, LAS-CLIP demonstrates a strong ability to maintain subject fidelity in color, texture, and structural detail. In the neuron illustration, both the baseline and Alpha-CLIP produce outputs with less structural similarity to the original diagram, while LAS-CLIP preserves more of the branching structure. Similarly, in the spider web example, Alpha-CLIP introduces text-like artifacts, whereas LAS-CLIP retains both the spider and the web pattern. These qualitative observations suggest that the frozen encoder helps preserve visual representations for downstream generative tasks, consistent with the analysis in the main text.

\section{Mask Robustness Analysis}
\label{sec:appendix_robustness}

To quantitatively evaluate robustness to imperfect masks, we systematically perturb the ground-truth segmentation masks on ImageNet-S using three strategies. These strategies include erosion for under-segmentation, dilation for over-segmentation, and spatial shifts. Each perturbation is calibrated to produce masks at approximate IoU levels of 0.90, 0.70, 0.50, and 0.30 relative to the ground truth. All experiments use the ViT-L/14 backbone.

\begin{figure}[h]
  \centering
  \includegraphics[width=\linewidth]{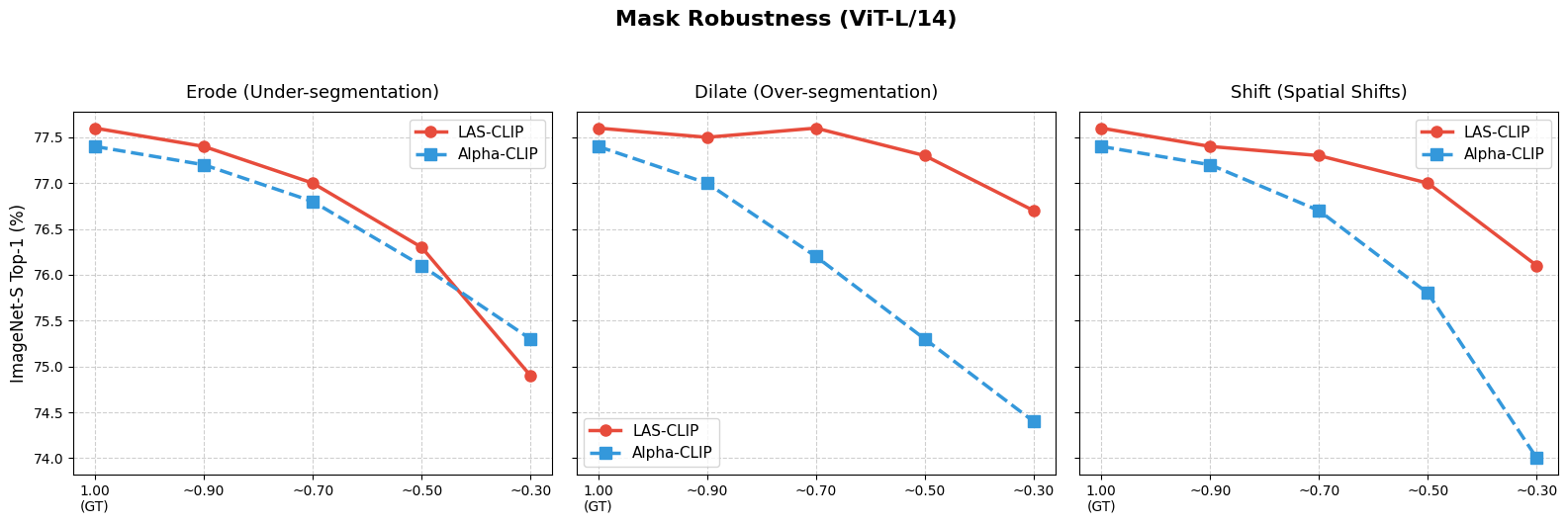}
  \caption{Mask robustness comparison on ImageNet-S (ViT-L/14). LAS-CLIP degrades more gracefully than Alpha-CLIP under dilation and spatial shift perturbations.}
  \label{fig:robustness}
\end{figure}

Figure~\ref{fig:robustness} reports the results. Under erosion, both methods degrade comparably. Under dilation and spatial shifts, LAS-CLIP exhibits substantially smaller accuracy drops than Alpha-CLIP~\cite{alphaclip}, especially at lower IoU levels. At IoU~$\approx$~0.30, the gap reaches 2.0\% for dilation and 1.7\% for spatial shifts. These results indicate that LAS-CLIP's token-conditioned gating mechanism mitigates over-reliance on the spatial mask prior, consistent with the qualitative attention analysis in the main paper.

\section{Training Data Source Ablation}
\label{sec:appendix_data_source}

To address whether the performance gains of LAS-CLIP stem from the adapter design or from the choice of mask generation pipeline, we retrain LAS-CLIP using SAM-generated masks on the same 100K training samples and compare against the default YOLOE-based pipeline.

Table~\ref{tab:data_cls} reports ImageNet-S classification results. LAS-CLIP trained with SAM masks consistently outperforms both LAS-CLIP with YOLOE masks and Alpha-CLIP~\cite{alphaclip} across both backbones, confirming that the adapter architecture is the primary source of improvement rather than the data pipeline.

Table~\ref{tab:data_rec} reports RefCOCO results. The SAM-trained and YOLOE-trained variants perform comparably, with YOLOE slightly ahead on most splits despite its noisier masks. This indicates that LAS-CLIP is robust to mask quality and that the observed improvements over Alpha-CLIP are not an artifact of the mask source.

\begin{table}[h]
\centering
\caption{Training data source ablation on ImageNet-S. Top-1/Top-5 accuracy (\%). Best in \textbf{bold}.}
\label{tab:data_cls}
\setlength{\tabcolsep}{6pt}
\begin{tabular}{l cc cc}
\toprule
\multirow{2}{*}{Method / Mask Source} & \multicolumn{2}{c}{ViT-B/16} & \multicolumn{2}{c}{ViT-L/14} \\
\cmidrule(lr){2-3} \cmidrule(lr){4-5}
 & Top-1 & Top-5 & Top-1 & Top-5 \\
\midrule
Alpha-CLIP (SAM)  & 68.9 & 90.5 & 77.4 & 94.5 \\
LAS-CLIP (YOLOE)  & 69.9 & 90.9 & 77.6 & 94.0 \\
LAS-CLIP (SAM)    & \textbf{70.8} & \textbf{91.3} & \textbf{78.1} & \textbf{94.1} \\
\bottomrule
\end{tabular}
\end{table}

\begin{table}[h]
\centering
\caption{Training data source ablation on RefCOCO / RefCOCO+ / RefCOCOg. Accuracy (\%). Best in \textbf{bold}.}
\label{tab:data_rec}
\setlength{\tabcolsep}{3pt}
\begin{tabular}{l ccc ccc cc}
\toprule
\multirow{2}{*}{Method / Source} & \multicolumn{3}{c}{RefCOCO} & \multicolumn{3}{c}{RefCOCO+} & \multicolumn{2}{c}{RefCOCOg} \\
\cmidrule(lr){2-4} \cmidrule(lr){5-7} \cmidrule(lr){8-9}
 & val & testA & testB & val & testA & testB & val & test \\
\midrule
Alpha-CLIP (SAM)  & 55.7 & 61.1 & \textbf{50.3} & 55.6 & 62.7 & 46.4 & 61.2 & \textbf{62.0} \\
LAS-CLIP (YOLOE)  & \textbf{57.5} & \textbf{64.1} & \textbf{50.3} & \textbf{57.4} & \textbf{66.0} & 46.6 & \textbf{61.7} & 61.7 \\
LAS-CLIP (SAM)    & 56.7 & 64.0 & 50.1 & 56.9 & 65.8 & \textbf{46.7} & 61.2 & 61.2 \\
\bottomrule
\end{tabular}
\end{table}

\section{Inference Efficiency}
\label{sec:appendix_efficiency}

Table~\ref{tab:efficiency} benchmarks inference efficiency on a single NVIDIA T4 GPU using the ViT-L/14 backbone. LAS-CLIP achieves throughput and latency comparable to Alpha-CLIP across all batch sizes, with marginal differences in FPS and per-image latency. The GFLOPs overhead is negligible at 0.3\% over vanilla CLIP. The primary trade-off is GPU memory. LAS-CLIP consumes approximately 2.6$\times$ the memory of vanilla CLIP at batch size 64, compared to 1.8$\times$ for Alpha-CLIP. This higher footprint arises from storing dense $N \times N$ attention bias matrices for each head and adapted layer. For applications where memory is constrained, reducing the number of adapter layers $K$ or lowering the adapter hidden dimension $h$ can mitigate this cost.

\begin{table}[h]
\centering
\caption{Inference efficiency on ViT-L/14 (single T4 GPU). FPS, latency, memory, and GFLOPs are reported across batch sizes.}
\label{tab:efficiency}
\setlength{\tabcolsep}{4pt}
\begin{tabular}{l c rr rr r r}
\toprule
Method & Batch & FPS & $\pm$std & Lat. (ms) & $\pm$std & Mem (MB) & GFLOPs \\
\midrule
CLIP       & 1  & 70.7 & 3.1 & 14.2 & 0.6 & 974  & 81.01 \\
CLIP       & 4  & 85.2 & 0.3 & 46.9 & 0.2 & 1000 & 20.25 \\
CLIP       & 16 & 80.4 & 1.1 & 199.1 & 2.8 & 1096 & 5.06 \\
CLIP       & 64 & 82.2 & 1.5 & 778.5 & 14.1 & 1480 & 1.27 \\
\midrule
Alpha-CLIP & 1  & 58.8 & 0.4 & 17.0 & 0.1 & 1925 & 81.06 \\
Alpha-CLIP & 4  & 65.6 & 0.4 & 61.0 & 0.4 & 1959 & 20.27 \\
Alpha-CLIP & 16 & 59.3 & 1.1 & 270.1 & 5.1 & 2084 & 5.07 \\
Alpha-CLIP & 64 & 64.3 & 1.1 & 995.3 & 17.1 & 2591 & 1.27 \\
\midrule
LAS-CLIP   & 1  & 61.7 & 0.4 & 16.2 & 0.1 & 2873 & 81.32 \\
LAS-CLIP   & 4  & 68.1 & 1.1 & 58.7 & 1.0 & 2928 & 20.33 \\
LAS-CLIP   & 16 & 61.5 & 0.4 & 260.4 & 1.9 & 3143 & 5.08 \\
LAS-CLIP   & 64 & 69.3 & 1.3 & 923.3 & 17.9 & 3915 & 1.27 \\
\bottomrule
\end{tabular}
\end{table}

\FloatBarrier
\bibliographystyle{splncs04}
\bibliography{main}
\end{document}